\documentclass[sigconf]{acmart}
\AtBeginDocument{%
  }
\usepackage{balance}

\usepackage{makecell}
\usepackage{multirow}
\usepackage{CJKutf8}
\usepackage{enumitem}
\usepackage{needspace}
\copyrightyear{2025}
\acmYear{2025}
\setcopyright{acmlicensed}\acmConference[MM '25]{Proceedings of the 33rd ACM International Conference on Multimedia}{October 27--31, 2025}{Dublin, Ireland}
\acmBooktitle{Proceedings of the 33rd ACM International Conference on Multimedia (MM '25), October 27--31, 2025, Dublin, Ireland}
\acmDOI{10.1145/3746027.3755497}
\acmISBN{979-8-4007-2035-2/2025/10}

\begin{document}

\title{Dual Enhancement on 3D Vision-Language Perception for Monocular 3D Visual Grounding}

\author{Yuzhen Li}
\orcid{0009-0000-3540-0627}
\affiliation{%
  \institution{School of Artificial Intelligence and Robotics, Hunan University}
  \city{Changsha}
  \state{Hunan}
  \country{China}
}
\email{zzrs@hnu.edu.cn}

\author{Min Liu}
\authornote{Corresponding author}
\orcid{0000-0001-6406-4896}
\affiliation{%
  \institution{School of Artificial Intelligence and Robotics, Hunan University}
  \city{Changsha}
  \state{Hunan}
  \country{China}
}
\email{liu_min@hnu.edu.cn}

\author{Yuan Bian}
\orcid{0000-0003-3995-4402}
\affiliation{%
  \institution{School of Artificial Intelligence and Robotics, Hunan University}
  \city{Changsha}
  \state{Hunan}
  \country{China}
}
\email{yuanbian@hnu.edu.cn}

\author{Xueping Wang}
\orcid{0000-0003-4862-8975}
\affiliation{%
  \institution{College of Information Science and Engineering, Hunan Normal University}
  \city{Changsha}
  \state{Hunan}
  \country{China}
}
\email{wang_xueping@hnu.edu.cn}

\author{Zhaoyang Li}
\orcid{0000-0003-4762-2993}
\affiliation{%
  \institution{School of Artificial Intelligence and Robotics, Hunan University}
  \city{Changsha}
  \state{Hunan}
  \country{China}
}
\email{zhaoyli@hnu.edu.cn}

\author{Gen Li}
\orcid{0000-0001-6636-1106}
\affiliation{%
  \institution{School of Informatics, University of Edinburgh}
  \city{Edinburgh}
  \country{United Kingdom}
}
\email{li.gen@ed.ac.uk}

\author{Yaonan Wang}
\orcid{0009-0004-5365-6254}
\affiliation{%
  \institution{School of Artificial Intelligence and Robotics, Hunan University}
  \city{Changsha}
  \state{Hunan}
  \country{China}
}
\email{yaonan@hnu.edu.cn}
\renewcommand{\shortauthors}{Yuzhen Li et al.}

\begin{abstract}
Monocular 3D visual grounding is a novel task that aims to locate 3D objects in RGB images using text descriptions with explicit geometry information. Despite the inclusion of geometry details in the text, we observe that the text embeddings are sensitive to the magnitude of numerical values but largely ignore the associated measurement units. For example, simply equidistant mapping the length with unit `meters’ to `decimeters’ or `centimeters’ leads to severe performance degradation, even though the physical length remains equivalent. This observation signifies the weak 3D comprehension of pre-trained language model, which generates misguiding text features to hinder 3D perception. Therefore, we propose to enhance the 3D perception of model on text embeddings and geometry features with two simple and effective methods. Firstly, we introduce a pre-processing method named 3D-text Enhancement (3DTE), which enhances the comprehension of mapping relationships between different units by augmenting the diversity of distance descriptors in text queries. Next, we propose a Text-Guided Geometry Enhancement (TGE) module to further enhance the 3D-text information by projecting the basic text features into geometrically consistent space. These 3D-enhanced text features are then leveraged to precisely guide the attention of geometry features. We evaluate the proposed method through extensive comparisons and ablation studies on the Mono3DRefer dataset. Experimental results demonstrate substantial improvements over previous methods, achieving new state-of-the-art results with a notable accuracy gain of 11.94\% in the `Far' scenario. Our code will be made publicly available.
\end{abstract}

\ccsdesc[500]{Information systems~Multimedia and multimodal retrieval}
\ccsdesc[300]{Computing methodologies~Object detection}

\keywords{Monocular 3D Visual Grounding, Multi-model Learning, 3D-text}


\maketitle

\section{Introduction}
Language-driven object understanding constitutes a fundamental capability for intelligent systems and human-robot interaction in physical environments. 2D visual grounding methodologies \cite{ruilinyao24,linhuixiao24,MinghangZheng24,shenxinchen24,chenchenjing20,dengj21,yangl22,zhany23} have achieved remarkable progress in image-based scene interpretation, but their inherent dimensional constraints prevent accurate spatial comprehension of real-world objects. This limitation has driven researchers to explore multi-modal solutions through feature fusion technologies. Current approaches primarily employ RGB-D scanning systems \cite{zehantan24,chendz20,achlioptasp20} for indoor scene reconstruction or adopt LiDAR-camera configurations \cite{linz25,jinyuanli25,shutinghe24} for outdoor robotic applications. However, the widespread deployment of these methodologies remains constrained by the high costs and limitations associated with RGB-D and LiDAR scanning devices.

\begin{figure*}[htbp]
  \includegraphics[width=\textwidth]{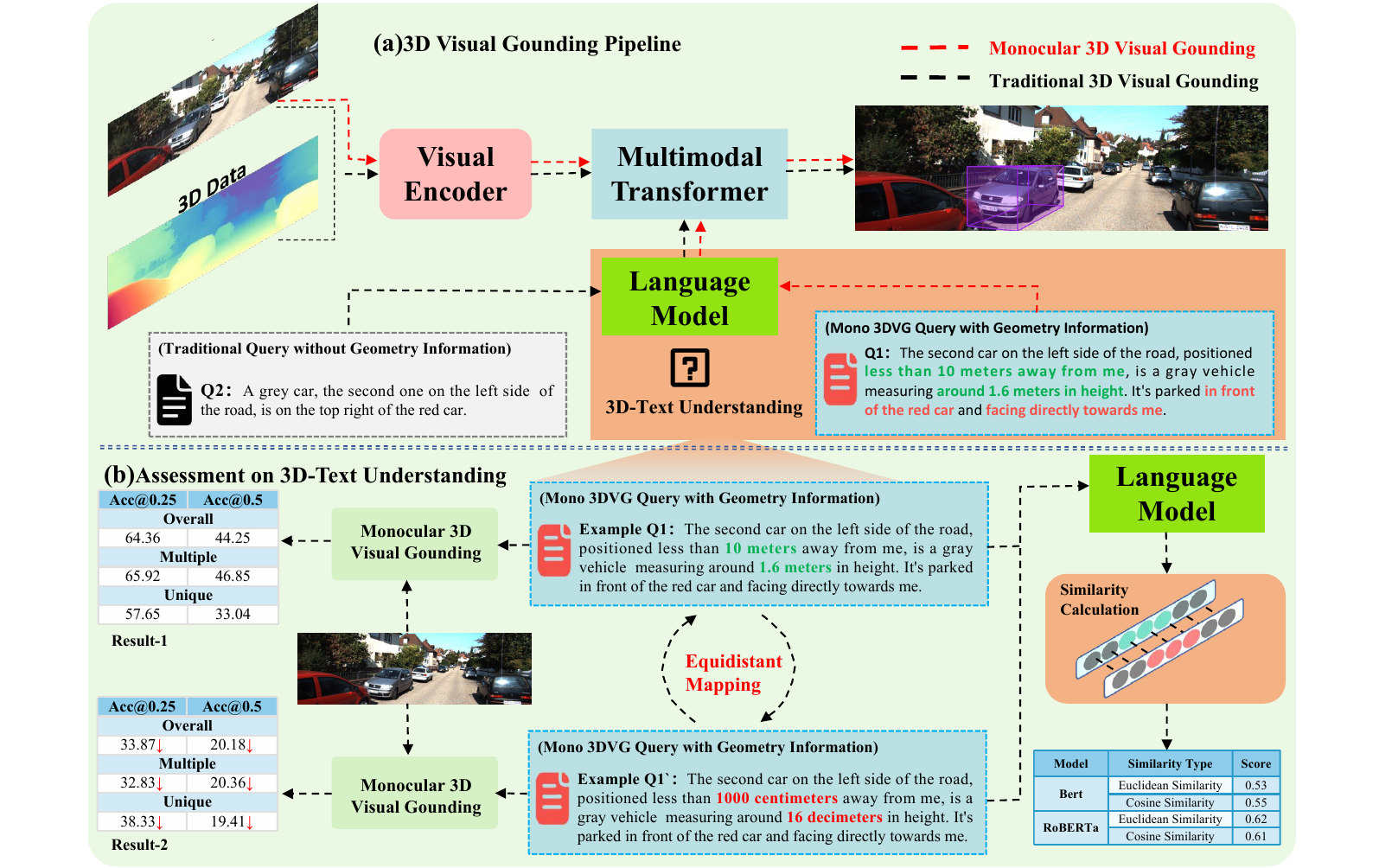}
  \caption{(a) This figure illustrates the differences between Monocular 3D Visual Grounding (red dashed lines) and Traditional 3D Visual Grounding (black dashed lines) in terms of input data; (b) Quantitative analysis of 3D-text understanding for language models. We compare the results of three scenarios (Overall, Multiple and Unique) and the feature similarity between all original queries and their counterparts after depth-wise equidistant mapping (e.g., converting `10 meters’ to `1000 centimeters’).}
  \label{fig:1}
\end{figure*}

Monocular 3D object detection \cite{huangk22,brazilg23,zli24} is capable of obtaining 3D spatial localization of objects using a single-view RGB image,  while overlooking semantic comprehension of 3D environments. This limitation also restricts their capacity to interpret human linguistic instructions for task-specific object localization. To address this critical challenge in human-machine collaboration systems (e.g., drones, autonomous vehicles, and robotic platforms), the paper \cite{zhany24} has introduced a novel task — monocular 3D visual grounding, which aims at localizing 3D objects in a single RGB image using language descriptions with geometry information. The fundamental difference between traditional 3D visual grounding and monocular 3D visual grounding is illustrated in Figure~\ref{fig:1} (a). Specifically, the monocular approach requires only RGB images as visual input (denoted by red dashed lines), whereas queries must be supplemented with explicit geometry descriptions. For clarity in the following paper, the three-dimension-related text (like 10 meters) within the captions will be formally designated as `3D-text' (3DT).

The monocular 3D visual grounding task supplements textual queries with geometry descriptions to compensate for depth ambiguity. However, through several comparative experiments, we observe that language models (BERT and RoBERTa) commonly employed in visual grounding tasks exhibit limited 3D-text perceptual capability, as depicted in Figure~\ref{fig:1} (b). We firstly apply a random equidistant mapping operation to all original queries Q1, generating transformed queries Q1'. This process maintains the syntactic structure and semantic information of the queries unchanged. Given that all Q1 and Q1' reflect identical semantic content and sentence structure, they should theoretically yield comparable results when used as textual inputs for monocular 3D visual grounding. However, as shown in the left part of Figure~\ref{fig:1} (b), the average accuracy of result-2 experiences significant degradation when all Q1' queries are employed as inputs. We only present three scenarios due to space constraints, though experimental results indicate a substantial decline across all scenarios. Meanwhile, with the high consistency between all Q1 and Q1', their corresponding encoded features from same layer of the text branch are supposed to exhibit strong similarities. Nevertheless, as shown in the right panel of Figure~\ref{fig:1} (b), the low average scores for both cosine similarity and euclidean similarity indicate major discrepancies. The severe performance degradation and low similarity scores reveal that the model struggles to recognize subtle changes in physical distances when expressed in different unites, suggesting a limited understanding of 3D-related textual information.
In addition to the weak 3D perception of language models, the conventional depth encoder exhibits insufficient interaction between geometric and textual features, while also lacking a specialized mapping layer to produce suitable text features optimally for interaction with geometry features.

In light of the above limitations, we build a framework to enhance the 3D perception capabilities of baseline network. First of all, we introduce 3D-text Enhancement (3DTE), an effective pre-processing method designed to enhance the diversity of depth-related units in captions, such as depth, height, and length, without altering the sentence's meaning and structure.
By employing this equidistant random mapping, it compels language models to develop a robust understanding of conversion relationships between depth values and units. Subsequently, we propose a Text-Guided Geometry Enhancement (TGE) module comprising two sequential operations to refine geometry features. This enhancement is crucial as geometry features derived from depth encoder provide essential depth cues for object positioning. Initially, the basic text features are projected into a geometrically consistent latent space that bridges basic text features and 3D visual features, thereby generating the 3D-enhanced text features. This 3D-enhanced text feature subsequently interacts with geometry features through a multimodal cross-attention layer, effectively fusing rich 3D details with geometry features. Through the aforementioned text-guided refinement process, it generates optimized geometry features enriched with 3D-specific details, thereby significantly improving 3D localization accuracy.

\raggedbottom
To summarize, we contribute to the research on monocular 3D visual grounding in the following aspects:
\begin{itemize}
\item[$\bullet$] We identify a previously overlooked limitation of widely used pre-trained language models in monocular 3D visual grounding, revealing their weak 3D-text perception capability.
\item[$\bullet$] We introduce a pre-processing method to augment the representational diversity of depth-related units, implicitly facilitating the model's understanding of equidistant mapping for monocular 3D visual grounding.
\item[$\bullet$] We propose a Text-Guided Geometry Enhancement (TGE) Module to generate 3D-enhanced text features by a projection layer and refine geometry features by a multimodal cross-attention layer.
\item[$\bullet$] We conduct extensive experiments to validate the merits of our method, and the results demonstrate that our method significantly outperforms all baselines and achieves a maximum accuracy improvement with 11.94\% on the `Far’ scenario.
\end{itemize}

\section{RELATED WORK}
\subsection{Monocular 3D Object Detection}
Monocular 3D object detection, which aims to accurately predict 3D bounding boxes, can be categorized into two main groups based on whether additional data beyond a single image is utilized. Methods relying solely on a single image such as M3D-RPN \cite{brazilg19} , which employs a standalone 3D region proposal network and proposes a depth-wise convolution. SMOKE \cite{liuz20} and FCOS3D \cite{wangt21} predict key-points to estimate the size and position of 3D bounding boxes based on a CenterNet \cite{zhoux19} framework. MonoLss \cite{zli24} introduces a novel Learnable Sample Selection (LSS) module designed to adaptively select samples, addressing the issue that not all features are equally effective for learning 3D properties. MonoFlex \cite{zhangy21} enhances truncated obstacle prediction with an edge heat-map and corresponding fusion module. MonoPair \cite{yongjianchen20} exploring pairwise spatial relationships to improve prediction accuracy, particularly for occluded objects. MonoEF \cite{yunsongzhou21} predicts camera extrinsic parameters via vanishing points and horizon detection, and rectifies perturbative features in latent space by a converter. MonoCon \cite{tianfuwu22} learns auxiliary monocular contexts mapping from the 3D bounding boxes during training and removes the auxiliary branch for higher efficiency in inference. MonoDDE \cite{zhuolingli22} exploits depth cues in monocular image and designs a model to produce 20 depths for each target. The second kind of research direction incorporates extra data, such as depth maps, point clouds, and CAD models to enrich visual perception for 3D detection. D4LCN \cite{mingyuding20} introduces depth-guided convolution  that dynamically adjust receptive fields based on estimated depth information. DID-M3D \cite{liangpeng22} further disentangles instance depths into attribute depths and visual depths through dense depth map analysis. ROI-10D \cite{fabianm19} regresses 3D bounding box by estimating dense depth maps. For LIDAR-based strategies, CaDDN \cite{codyreading21} employs a monocular network to estimate depth maps generated by LIDAR data, subsequently transforming features into bird's-eye view representations for prediction. CMKD \cite{yuhong22} enhances this paradigm through cross-modal knowledge distillation from LIDAR to visual modalities. In addition to depth maps and LIDAR, AutoShape \cite{zongdailiu21} leverages CAD model-derived keypoint generation to overcome sparse constraints.

\subsection{2D Visual Grounding}
Visual grounding, a task evolved from object detection, aims to establish precise alignment between linguistic descriptions and corresponding regions in visual content. Early approaches in this field predominantly adopted two-stage frameworks \cite{daqingliu19,sibeiyang19,lichengyu18,hanwangzhang18}, which decouple the process into sequential steps: proposal generation and cross-modal matching. In the first stage, an object detector (e.g., Faster R-CNN) generates region proposals independently of textual input, potentially introducing semantic misalignment due to the lack of linguistic guidance. In the second stage, a ranking network computes similarity scores between text features and candidate regions using losses like the maximum-margin ranking loss \cite{lichengyu18} or binary classification loss \cite{plummerba18}, selecting the highest-scoring region as the grounding result. To address the semantic gap between modalities, MattNet \cite{lichengyu18} decomposes text into structural components (subject, location, relationships) for fine-grained visual-linguistic reasoning. Conversely, proposal-free approaches \cite{xinpengchen18,yueliao20,zhenyuanyang19} implement dense spatial feature integration, enabling direct localization prediction through comprehensive cross-modal analysis. A representative implementation is FAOA \cite{zhenyuanyang19}, which achieves end-to-end localization by integrating concatenated multi-modal features into a YOLOv3 detection framework \cite{yolov3}. To handle linguistically complex expressions, ReSC \cite{yangz20} introduces a recursive sub-query decomposition strategy, progressively refining visual-textual alignment through iterative attention mechanisms over constituent query elements.

Given the successful application of transformers in various tasks, a series of transformer - based networks have been proposed for application in visual grounding tasks. Transformer-based approach is first introduced by TransVG \cite{dengj21}. Driven by the inherent flexibility of transformer architectures in processing cross-modal interactions, subsequent research has extended this paradigm through feature refinement and context-aware representation learning. Referring Transformer \cite{zhuolingli22} utilize contextual phrase queries for simultaneous region identification and segmentation with a single model. MDETR \cite{kamatha21} demonstrates end-to-end localization capabilities using a transformer-based encoder-decoder framework with aligned feature as input. However, these studies are unable to acquire the 3D coordinates of objects in the real - world, significantly restricting their practical application.

\subsection{3D Visual Grounding}
To facilitate the development of 3D visual grounding,  Referit3D \cite{achlioptasp20} and ScanRefer \cite{chendz20} initial build the benchmark datasets. Early approaches emulated 2D visual grounding tasks by utilizing the two-stage strategy. Like PointNet++ \cite{qic17}, it utilizes a pre-trained detector for object proposals and feature extraction. To augment semantic information, SAT \cite{yangz2021} integrates 2D objects semantics to enhance model training. To tackle the challenges associated with interpreting complex descriptions, making sense of point cloud scenes, and pinpointing target objects, Feng et al. \cite{fengm21} introduce three dedicated modules: a language scene graph module, a multi-level 3D proposal relation graph module, and a description - guided 3D visual graph module. 3DVG-Trans \cite{zhaol21}, TransRefer3D \cite{hed21}, Multi-View Trans \cite{huangs22}, and LanguageRefer \cite{rohj22} all develop architectures based on transformer. Unified frameworks \cite{chendz22,caid22} are designed to handle both dense captioning and visual grounding tasks. A new task involving 3D visual grounding within RGB-D images is introduced by Liu et al \cite{liuh21}. Previous studies have mainly focused on indoor conditions, with furniture serving as the object of interest. To expand the application scope, Lin et al. \cite{linz25} introduced the task to large-scale dynamic outdoor scenes, using 2D images and 3D point clouds. However, the high cost and limited accessibility of LiDAR and RGB-D devices for capturing visual data restrict their extensive application. Mono3DVG \cite{zhany24} proposes a new task that captions enriched by geometry details combine with a single RGB image to predict the 3D bounding box, named monocular 3D visual grounding.

\section{Method}
\subsection{Preliminary}
As our approach builds upon the baseline framework \cite{zhany24}, we briefly review its architecture before presenting our methods. RoBERTa-base \cite{liuy19} paired with a linear layer is employed to embed basic text features $f_t$, and ResNet-50 \cite{hek16} combined with a linear layer is utilized to derive multi-scale visual features $f_v$. Following Zhang et al. \cite{zhangr22}, a lightweight depth predictor is adopted to extract geometry features $f_g$. We then follow the two-branch encoder comprising: (1) a depth encoder (equation 1) implemented with a single transformer layer to encode geometry embeddings, and (2) a visual encoder (equation 2) that employs multi-scale deformable attention (MSDA) for efficient spatial modeling, coupled with a cross-attention layer that injects textual guidance before the feed-forward network. The following adapter is build for processing geometry and visual features through text-guided cross-attention. The decoder employs progressive attention refinement, where a learnable query dynamically integrates geometry priors, text, and multi-scale visual features through sequential cross-modal fusion stages. Further details can be found in \cite{zhany24}.

\begin{equation}
p_g=FFN(MHSA(f_g)),
\end{equation}

\begin{equation}
p_v=FFN(MHCA(MSDA(f_v),f_t)).
\end{equation}

\subsection{3D-Text Enhancement}
As illustrated by the Q1 query example in Figure~\ref{fig:1} (a), captions in the Mono3DRefer dataset contain abundant geometry distance descriptors (highlighted in green) that all use "meters" as measurement units. This lexical redundancy restricts language models' comprehension on mapping relations of different depth units and induces insensitivity in three-dimensional information. Inspired by the linguistic augmentation strategy proposed in \cite{jwei19}, we develop a novel method named 3D-text Enhancement (3DTE) that enhances depth-related text diversity while strictly preserving  text-semantic consistency.

\begin{figure}[h]
  \centering
  \includegraphics[width=\linewidth]{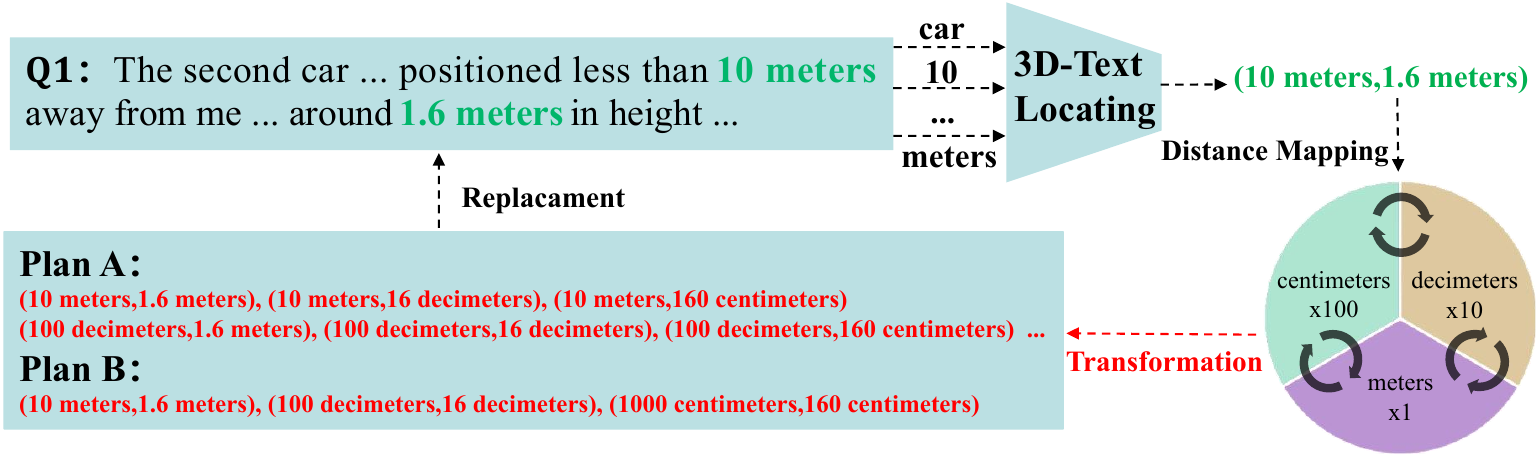}
  \caption{An Example of 3D-text Enhancement. Plan A implements per-unit random replacement within Query 1 (Q1), while Plan B employs equidistant replacement but maintains holistic units as same across all Q1 distance parameters.}
  \label{fig:2}
\end{figure}

\begin{figure*}[htbp]
  \center{\includegraphics[width=\textwidth]{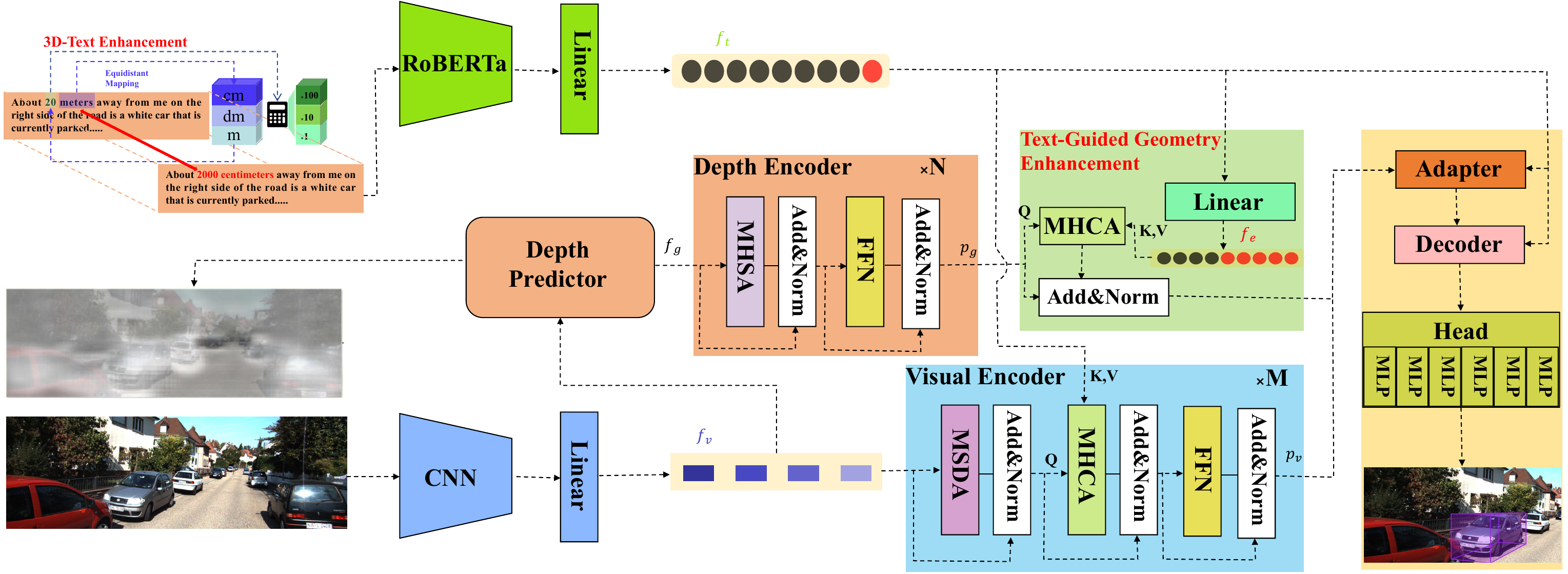}}
  \caption{Overview of the framework. The RoBERTa model, visual encoder and depth encoder aim to extract textual, visual, and geometric features. The Text-Guided Enhancement module is proposed to enhance the extraction of 3D-text information and refinement of geometry features. The adapter refines the referred object features on vision and geometry. The decoder and head with multiple MLPS are build for the 2D and 3D attributes prediction.}
  \label{fig:33}
\end{figure*}

3DTE introduces randomly-equidistant mapping that maintains the same semantic information through unit conversion. In practical visual grounding applications where object scales predominantly operate within meter-to-centimeter ranges, we strategically constrain unit substitution in textual descriptions to three hierarchical levels: meter, decimeter, and centimeter. During the data augmentation process, we assign equal probabilities to the three depth units (meters, decimeters, and centimeters). In our design, we deliberately select the approach of independently mapping each distance unit within sentences (Plan A, Figure~\ref{fig:2}) over maintaining same unit throughout the text (Plan B, Figure~\ref{fig:2}). This choice to Plan A stems from our hypothesis that per-unit augmentation would better enhance textual diversity, which is also validated by the experimental results presented in Section 4. As shown on Figure~\ref{fig:2}, the method begins with locating 3D-text in the original text Q1. Subsequently, we perform equidistant mapping of depth, assigning equal transition probabilities across meter, decimeter, and centimeter units. The final augmented text Q1' is generated through replacement of original 3D-text with these remapped 3D descriptors.

\subsection{Assessment on Language Model}
The standard pipeline for monocular 3D Visual Grounding (3DVG) is illustrated in Figure~\ref{fig:1} (a). It is worth noting that the textual descriptions in monocular 3DVG datasets are rich in 3D-text information. This characteristic requires language models to accurately parse 3D-text features from description. Otherwise, it will lead to a misalignment of multi-modal features, ultimately degrading localization precision. Hence, we design a comparative experiment to verify the 3D-text understanding capabilities of two widely used language models, RoBERTa and BERT, in visual grounding tasks.

We firstly transform the original textual descriptions Q1 into augmented versions Q1' using the method from Section 3.2, with a strategic refinement: constraining the mapping transformation exclusively to decimeter and centimeter units. This modification is specifically designed to maximize difference between Q1 and Q1' in distance unit representation. The original text Q1 and augmented text Q1' are subsequently encoded by a language model to obtain their respective feature representations $f_O$ and $f_A$. We then compute both euclidean similarity $S_E$ and cosine similarity $S_C$ between these representations using Equation (3) and Equation (4). Let $m$ denote the text mask in the formula. To compute Euclidean similarity, we first apply element-wise product between the text embeddings and mask m to eliminate interference from padding tokens. The masked embeddings then undergo euclidean distance calculation, followed by averaging operation. The final similarity score is obtained by subtracting from 1. For cosine similarity, we directly calculate the cosine distance between the masked text embeddings after the same element-wise masking operation.

\begin{equation}
S_E=1-\mathrm{mean}(\sqrt{\parallel(f_0-f_A)\cdot m\parallel^2}),
\end{equation}

\begin{equation}
S_{C}=\operatorname{mean}\left(\frac{f_{0} \cdot f_{A}\cdot m}{\left\|f_{0}\right\|\cdot\left\|f_{A}\right\| \cdot m}\right).
\end{equation}

The magnitudes of euclidean similarity and cosine similarity serve as indicators of the text encoder's comprehension capability for 3D-text representations. Although a equidistant mapping transformation has been applied, the underlying meaning and the 3D details remain consistent. Therefore, higher similarity scores demonstrate better 3D information understanding, as they reflect better preservation of 3D-semantic relationships in the embedding space. As illustrated in right-corner part  of Figure~\ref{fig:1} (b), the experimental results show that the RoBERTa and BERT exhibit weak 3D awareness, with minimal sensitivity to explicit 3D descriptions embedded in the text.

\subsection{Text-Guided Geometry Enhancement}
Based on the experimental analysis in Section 3.3, we conclude that the RoBERTa exhibits insensitivity to 3D-text information, which may result in the lack of precise 3D information guidance for geometry features. Therefore, we propose a Text-Guided Geometry Enhancement (TGE) module to enhance the extraction of three-dimensional information from basic text features and provide the enhanced 3D features for geometry encoding. The overview of our framework is illustrated in Figure~\ref{fig:33}. 

In the pursuit of enhancing the integration of 3D-text information into geometry features $p_g$, we propose a architecture that leverages multi-head cross-attention (MHCA) mechanisms. As depicted in Figure~\ref{fig:33}, our approach introduces a MHCA layer designed to infuse precise 3D-text guidance into the geometry features. This multimodal interaction is necessary to enhance geometry features for the deep understanding of depth relationships described within textual descriptions.

However, our initial quantitative analysis in Section 3.3 highlight a significant challenge: the ambiguous 3D information within the basic text features $f_t$. To mitigate this issue, we develop an auxiliary linear layer prior to the MHCA layer. This additional layer serves as a strengthen step, specifically tailored to enhance the extraction of 3D-text features from the input $f_t$. The output denoted as $f_e$ represents the enhanced 3D-text features. The features are then utilized as key-value pairs (k,v) within the MHCA layer. By doing so, we ensure that the MHCA layer receives accurate and enriched 3D textual cues, which are essential for the effective enhancement of geometry features. The simple fully-connected layer used in our approach demonstrates better understanding capabilities regarding different depth-unit mapping relationships, as demonstrated by subsequent experiments in section 4.

\begin{equation}
f_e=\sigma(W_p\cdot f_t+b_p),
\end{equation}

\begin{equation}
g_e=MHCA(q,k,v) ,q=p_{g},k=v=f_{e}.
\end{equation}

The equation 5 denotes a fully-connected layer, and the MHCA represents a multi-head cross-attention layer. The $W_p\in\mathbb{R}^{C\times C}$ in equation 5 denotes the learnable projection matrix and $\sigma$ represents the ReLU activation. This refinement enables injection of 3D-enhanced textual cues into geometry features through cross-modal attention, while maintaining computational efficiency via a standard fully-connected (FC) layer.

\subsection{Loss Function}
The final head employs multiple MLPS for 2D and 3D attribute prediction. Specifically, 2D properties encompass object classification, 2D box size, and the projected 3D center. Conversely, 3D parameters include 3D box size,  orientation, and depth values. The 2D loss is mathematically expressed as:

\begin{equation}
L_{2D}=\lambda_1L_{classs}+\lambda_2L_{lrtb}+\lambda_3L_{GIoU}+\lambda_4L_{xy3D},
\end{equation}where $\lambda_{1-4}$ is set to (2,5,2,10) following MonoDETR \cite{zhangr22}. $L_{class}$ is Focal loss \cite{linty17} for nine classes prediction. $L_{lrtb}$  and $L_{xy3D}$ adopt the L1 loss. $L_{GIoU}$ is the GIoU loss \cite{rezatofighih19} that regresses the 2D bounding boxes. The loss for 3D is defined as:

\begin{equation}
L_{3D}=L_{size3D}+L_{orien}+L_{depth}.
\end{equation}The terms $L_{size3D}$, $L_{orien}$ and $L_{depth}$ denote the 3D IoU oriented loss \cite{max21}, MultiBin loss and Laplacian aleatoric uncertainty loss \cite{chendz20}, respectively, employed to optimize the predicted 3D size, orientation, and depth parameters. Finally, a Focal loss is used to supervise the prediction of the depth map, denoted as $L_{dmap}$, and the overall loss is formulated as:

\begin{equation}
L_{overall}=L_{2D}+L_{3D}+L_{dmap}.
\end{equation}

\section{Experiments}
\subsection{Dataset}
The Mono3DRefer \cite{zhany24} dataset consists of 2,025 sampled images from the original KITTI dataset \cite{geiger2012we}, comprising 41,140 descriptions and a vocabulary size of 5,271 words. Except template-generated Sr3d dataset, Mono3DRefer has a comparable number of descriptions to ScanRefer and Nr3d. Comared to RGB-D and LiDAR visual annotations, Mono3DRefer establishes a maximum annotated object distance of 102m, while RGB-D and LiDAR sensors are limited to approximately 10m and 30m, respectively.

\subsection{Implementation Details}
Implementations utilize a GTX 3090 (24GiB) GPU with AdamW optimization (batch size=10, 60 epochs in total), initial learning rate $ 10^{-4}$ and weight decay $ 10^{-4}$. The learning rate decays by a factor of 10 after 40 epochs. A dropout rate of 0.1 is employed. Following Lin et al. \cite{linz25}, we evaluate the accuracy with 3D IoU threshold (0.25 and 0.5) as our metrics.

Similar to Mono3DVG \cite{zhany24}, we adopt the same baseline to guarantee fair comparisons. \textbf{Two-stage methods}: (1) CatRand randomly selects category-matched ground truth boxes as predictions to measure baseline difficulty; (2) (Cube RCNN \cite{brazilg23} + Rand) employs random selection from proposals generated by Cube RCNN [13] as prediction result; (3) (Cube RCNN \cite{brazilg23} + Best) determines the upper bound on how well the two-stage approaches work for Mono3DVG task by selecting best matched label box with proposals. \textbf{One-stage methods}: we adapt results of 2D visual grounding to 3D by back-projection, and we select four SOTA methods: ZSGNet \cite{sadhua19}, FAOA \cite{zhenyuanyang19}, ReSC \cite{yangs20}, and TransVG \cite{dengj21}. To investigate the impact of non-categorical information, we evaluate these baselines capabilities through three critical dimensions: (1) `\textbf{unique}’ and `\textbf{multiple}’ scenarios: the ’unique’ scenario comprises single-object scenarios, whereas the `multiple’ scenario encompasses multi-instance scenarios that contains multiple confused objects with the same category; (2) To quantify task difficulty, we evaluate metrics on depth intervals: \textbf{Near} (0-15m), \textbf{Medium} (15-35m), and \textbf{Far} (>35m); (3) To assess the influence of visual occlusion and truncation of objects, we also report metrics at varying levels of difficulty: \textbf{easy} cases maintain no occlusion and truncation ratios below 0.15, \textbf{moderate} cases involve no/partial occlusion and truncation under 0.3, and \textbf{hard} cases encompass severely obscured objects exceeding above thresholds.

\begin{table*}
  \caption{Comparisons with baselines. Results with underlines denote better performance than our bolded acc.}
  \label{tab:score1}
  \tabcolsep=0.020\linewidth
  \begin{tabular*}{\linewidth}{cccccccc}
    \Xhline{1pt}
    \multirow{2}{*}{Method}&\multirow{2}{*}{Type}&\multicolumn{2}{c}{Unique}&\multicolumn{2}{c}{Multiple}&\multicolumn{2}{c}{Overall}\\
    &&Acc@0.25&Acc@0.5&Acc@0.25&Acc@0.5&Acc@0.25&Acc@0.5\\
    \Xhline{0.5pt}
    CatRand	&Two-Stage&	\underline{100}&	\underline{100}	&24.47	&24.43	&38.69&	38.67 \\
    Cube R-CNN + Rand	&Two-Stage	&32.76	&14.61	&13.36	&7.21&	17.02&	8.60 \\
    Cube R-CNN + Best	&Two-Stage&	35.29	&16.67&	60.52	&32.99&	55.77	&29.92\\
    \Xhline{0.5pt}
    ZSGNet + backproj	&One-stage&	9.02&	0.29&	16.56	&2.23	&15.14&	1.87\\
    FAOA + backproj	&One-stage&	11.96	&2.06	&13.79	&2.12	&13.44&	2.11\\
    ReSC + backproj	&One-stage&	11.96&0.49&	23.69	&3.94&	21.48	&3.29\\
    \Xhline{0.5pt}
    TransVG + backproj&	Tran.-based&	15.78	&4.02	&21.84&	4.16	&20.70	&4.14\\
    Mono3DVG-TR	&Tran.-based&	57.65&	33.04	&65.92&	46.85	&64.36	&44.25\\
    \Xhline{0.5pt}
    \textbf{Mono3DVG-TGE(Ours)}	&Tran.-based	&\textbf{62.45}&	\textbf{44.22}&	\textbf{69.83}&	\textbf{52.83}&	\textbf{68.44}	&\textbf{51.21}\\
  \Xhline{1pt}
\end{tabular*}
\end{table*}

\begin{table*}
  \caption{Comparisons for near-medium-far scenarios and easy-moderate-hard scenarios. Results with underlines denote better performance than our bolded acc.}
  \label{tab:score2}
  \tabcolsep=0.013\linewidth
  \resizebox{\textwidth}{!}{
  \begin{tabular*}{\linewidth}{cccccccc}
    \Xhline{1pt}
    \multirow{2}{*}{Method}&\multirow{2}{*}{Type}&\multicolumn{2}{c}{Near/Easy}&\multicolumn{2}{c}{Medium/Moderate}&\multicolumn{2}{c}{Far/Hard}\\    &&Acc@0.25&Acc@0.5&Acc@0.25&Acc@0.5&Acc@0.25&Acc@0.5\\
    \Xhline{0.5pt}
    CatRand	&Two-Stage&	31.16/47.29	&31.05/47.26	&35.49/33.92	&35.49/33.92	&52.11/30.83	&\underline{52.11}/30.74 \\
    Cube R-CNN + Rand	&Two-Stage	&17.40/21.12	&11.45/11.41	&18.01/17.85	&8.15/8.01	&14.91/10.56	&6.38/5.18 \\
    Cube R-CNN + Best	&Two-Stage&	67.76/59.66&	41.45/33.05	&60.69/60.56&	30.35/33.45	&34.72/46.25	&17.01/22.52\\
    \Xhline{0.5pt}
    ZSGNet + backproj	&One-stage&	24.87/21.33&	0.59/3.35&	16.74/13.87	&3.71/0.63	&2.15/7.57	&0.07/0.84\\
    FAOA + backproj	&One-stage&	18.03/17.51	&0.53/3.43	&15.64/12.18&	3.95/1.34	&4.86/8.83&	0.62/0.90\\
    ReSC + backproj	&One-stage&	33.68/27.90&	0.59/5.71&	24.03/19.23&	6.15/1.97&4.24/14.41&	1.25/1.02\\
    \Xhline{0.5pt}
    TransVG + backproj&	Tran.-based&	29.34/28.88	&0.86/6.95	&25.05/16.41	&8.02/2.75&	4.17/12.91	&0.97/1.38\\
    Mono3DVG-TR	&Tran.-based&	64.74/72.36&	53.49/51.80	&75.44/69.23&	55.48/48.66&	45.07/49.01&	15.35/29.91\\
    \Xhline{0.5pt}
    \textbf{Mono3DVG-TGE(Ours)}	&Tran.-based	&\textbf{68.02}/\textbf{76.91}&	\textbf{56.64}/\textbf{60.30}&	\textbf{78.49}/\textbf{73.66}&	\textbf{61.87}/\textbf{56.48}&	\textbf{52.99}/\textbf{52.13}	&\textbf{27.29}/\textbf{33.99}\\
  \Xhline{1pt}
\end{tabular*}
}
\end{table*}


\subsection{Quantitative Comparisons and Analyses}
As shown in Table~\ref{tab:score1}, the CatRand model exhibits a performance disparity between the `unique' scenario and the 'multiple' scenario, achieving 100\% accuracy on the former versus merely 24\% on the latter. When an image contains a singular object instance (e.g., a single car), the provision of its categorical label (`car') proves adequate for unambiguous identification. Conversely, in scenarios involving multiple instances of the same category, supplementary contextual cues such as spatial relationships or distinct attributes must be integrated into the input to eliminate localization ambiguities. In Table~\ref{tab:score2}, CatRand achieves superior performance on the `far' scenario compared to other scenarios, while our method and other baseline approaches exhibit progressively degraded accuracy with increasing depth. This discrepancy can be attributed to the inherent characteristics of the `far' scenario, which contains fewer ambiguous objects. CatRand's strategy of random ground truth selection is effective in this scenario, whereas conventional methods face greater challenges in depth prediction. Our proposed model almost achieves the best performance across all scenarios, with exception of CatRand on `far'. In Table~\ref{tab:score2}, Our method achieves state-of-the-art performance across all scenarios, demonstrating significant improvements over baseline in Mono3DVG tasks. In particular, our method shows notable improvements under stricter evaluation criteria (acc@0.5), with accuracy gains of +11.94\%, +11.18\% and +8.5\% in the `far', `unique' and `easy' scenarios, respectively, even though our minimum improvement reaches an increase of +3.05\% in the medium scenario.

\begin{table}
  \caption{Text semantic similarity comparisons.}
  \label{tab:Understanding1}
  \begin{tabular}{ccc}
    \Xhline{1pt}
    Method&Type&Score\\
    \Xhline{0.5pt}
    \multirow{2}{*}{Mono3DVG-TR}&Euclidean similarity & 0.62 \\
    &Cosine similarity& 0.61 \\
    \Xhline{0.5pt}
    \multirow{2}{*}{Ours}&Euclidean similarity & 0.91 \\
    &Cosine similarity& 0.85 \\
  \Xhline{1pt}
\end{tabular}
\end{table}

As quantitatively demonstrated in Table~\ref{tab:Understanding1}, our model maintains consistent semantic representations (similarity >= 85\%) between the original and randomly equidistance-mapped 3D-enhanced features $f_e$. This robustness to comprehend distance mapping relations confirms our model's capacity to effectively preserve 3D-aware details, particularly in encoding critical 3D-text information that conventional approaches often ignore.

\begin{table}
  \caption{Performance comparisons on unseen depth unit.}
  \label{tab:milli}
  \begin{tabular}{ccccc}
    \Xhline{1pt}
    \multirow{2}{*}{Scenarios}&\multicolumn{2}{c}{Our}&\multicolumn{2}{c}{Mono3DVG-TR}\\
    &Acc@0.25&Acc@0.5&Acc@0.25&Acc@0.5\\
    \Xhline{0.5pt}
    Unique&60.29& 37.45&38.33& 19.41 \\
    \Xhline{0.5pt}
    Multiple&61.89 & 45.94&32.83& 20.36 \\
    \Xhline{0.5pt}
    Overall&61.59 & 44.34&33.87& 20.18 \\
    \Xhline{0.5pt}
    Near&64.93 & 54.47&43.75& 32.30 \\
    \Xhline{0.5pt}
    Medium&68.88 & 52.38&32.38& 18.90 \\
    \Xhline{0.5pt}
    Far&45.62 & 20.05&25.97& 9.58 \\
    \Xhline{0.5pt}
    Easy&71.85 & 54.29&43.13& 26.78 \\
    \Xhline{0.5pt}
    Moderate&63.52 & 46.76&32.68& 19.65 \\
    \Xhline{0.5pt}
    Hard&45.59 & 28.35&21.92& 11.41 \\
  \Xhline{1pt}
\end{tabular}
\end{table}

\begin{figure*}[htbp]
  \center{\includegraphics[width=\textwidth]{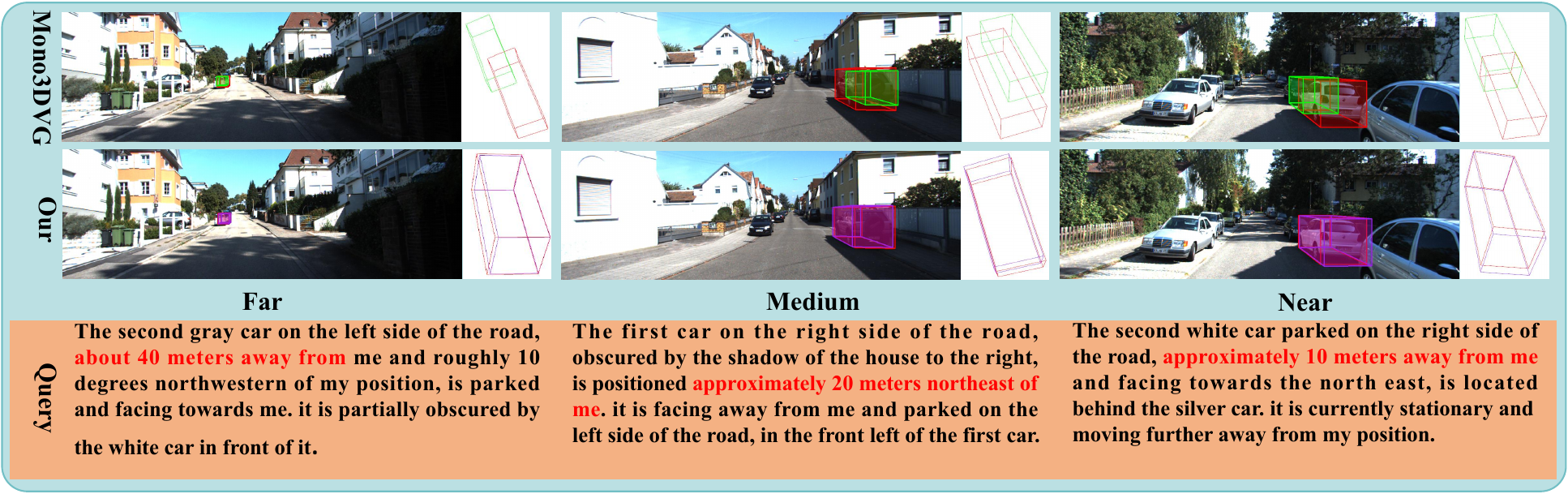}}
  \caption{Visualized results from Mono3DVG-TR and our proposed method. The red, purple and green boxes represent the ground truth, our predicted boxes and Mono3DVG-TR predicted boxes, respectively.}
  \label{fig:4}
\end{figure*}

The performance degradation shown in Figure~\ref{fig:1} (b) arises from the baseline model's limited ability to accurately perceive depth units that are not included in the training data. Therefore, we evaluate our proposed method under similar conditions. Since our proposed method is trained with text data on `meters', `decimeters' and `centimeters' , we map them to neighboring and unseen units: `kilometers' and `millimeter'. However, to prevent potential bias caused by extremely numerical representation (e.g., 1.8-meters-height mapping to 0.0018-kilometers-height), we map depth, height, and length descriptions in the test data to `millimeter' units. As shown in Table~\ref{tab:milli}, we compare the performance of our method and the baseline when processing text data with unseen distance units. Across all scenarios, our model maintains competitive performance compared to the original test data, whereas the baseline suffers a significant decline.

\begin{table}
  \caption{Comparisons on different 3D-text Enhancement Plans. Results with underlines denote better performance.}
  \label{tab:Understanding2}
  \begin{tabular}{ccccc}
    \Xhline{1pt}
    \multirow{2}{*}{Scenarios}&\multicolumn{2}{c}{Acc@0.25}&\multicolumn{2}{c}{Acc@0.5}\\
    &Plan A&Plan B&Plan A&Plan B\\
    \Xhline{0.5pt}
    Unique&\underline{62.22} & 61.33&\underline{43.63}&42.06 \\
    \Xhline{0.5pt}
    Multiple&\underline{69.08}&68.15&\underline{51.79}&51.74 \\
    \Xhline{0.5pt}
    Overall&\underline{68.16}&67.24&\underline{50.25}&49.92 \\
    \Xhline{0.5pt}
    Near&\underline{67.45}&66.72&55.71&\underline{56.04} \\
    \Xhline{0.5pt}
    Medium&\underline{78.45}&76.33&\underline{61.02}&60.98 \\
    \Xhline{0.5pt}
    Far&49.79&\underline{50.69}&\underline{25.07}&23.54 \\
    \Xhline{0.5pt}
    Easy&\underline{76.65}&76.57&\underline{59.79}&57.81 \\
    \Xhline{0.5pt}
    Moderate&\underline{71.55}&70.00&52.11&\underline{54.08} \\
     \Xhline{0.5pt}
    Hard&\underline{51.89}&51.53&33.32&33.32 \\
  \Xhline{1pt}
\end{tabular}
\end{table}

To quantitatively evaluate pre-processing strategies for descriptions, we conduct a comparative analysis between two proposed plans (Plan A and Plan B) as detailed in Section 3.2. The experimental results presented in Table~\ref{tab:Understanding2} compare their performance across nine distinct scenarios. Both plans show improvements over the baseline, with Plan A exhibiting statistically better performance across most evaluation metrics compared to Plan B.

\subsection{Qualitative Analysis}
Figure~\ref{fig:4} presents a comparative visualization of 3D grounding performance between the baseline Mono3DVG-TR and our proposed Mono3DVG-TGE. To establish clear visualization, we focus on no occlusion and no truncation vehicle instances across three representative distance ranges: near (<15m), medium (15-35m), and far (>35m) scenarios. From the qualitative comparisons, both methods exhibit similar capabilities in orientation estimation and 3D dimensional size prediction, while our approach demonstrates improved depth positioning capability. This improved depth awareness results in higher alignment between predicted and ground-truth 3D boxes, as reflected in the 3D box visualizations. These improvements validate the effectiveness of our Text-Guided Geometry Enhancement module.

\subsection{Ablation Studies}

\begin{table}
  \caption{The ablation studies of our proposed methods on the `Overall' scenario.}
  \label{tab:Understanding}
  \begin{tabular}{c|c|cc}
    \Xhline{1pt}
    3DTE&TGE&Acc@0.25&Acc@0.5\\
    \Xhline{0.5pt}
    -&- & 64.36&44.25 \\
    \checkmark &-& 68.16 &50.25\\
    -&\checkmark & 66.61 &49.29\\
    \checkmark&\checkmark & 68.44 &51.21\\
  \Xhline{1pt}
\end{tabular}
\end{table}

We perform ablation studies on the Mono3DRefer dataset to validate our 3D-text Enhancement (3DTE) pre-processing method and Text-Guided Geometry Enhancement (TGE) module, reporting their effectiveness through standard accuracy metrics at Acc@0.25 and Acc@0.5 on the overall scenario in Table~\ref{tab:Understanding}. Row 1 displays the baseline model's performance. Row 2 presents results using solely the 3DTE pre-processing method, and Row 3 demonstrates outcomes which only adds TGE module. Comparative analysis reveals that both proposed components independently achieve significant accuracy improvements (3DTE: +3.8/+6.0, TGE: +2.42/+4.28), validating their distinct contributions to the monocular 3D visual grounding task. The last row shows the integration of both components (3DTE+TGE), achieving a synergistic rise that surpasses the individual gains. 

\section{Conclusion}
This paper alleviates two critical limitations in current monocular 3D visual grounding architecture based on our observation. Our comparative experiments reveal the limitations of the widely used language models in 3D-text perception, as evidenced by low 3D-text feature similarity scores and performance degradation. Firstly, we resolve the linguistic redundancy through our 3D-text Enhancement (3DTE) method, which improves unit diversity while preserving semantic consistency, thereby implicitly strengthening language encoders' understanding of depth-value conversions. Secondly, the proposed Text-Guided Geometry Enhancement (TGE) module successfully refines the geometry features based on projected 3D-enhanced text features, providing more accurate depth information for 3D localization. Through comprehensive comparisons and ablation experiments, our method demonstrates state-of-the-art performance on most scenarios by overcoming the inherent limitations of conventional architectures. 

\section{Acknowledgements}
This work was supported in part by the National Natural Science Foundation of China under Grant 62221002, 62425305 and U22B205, in part by the Science and Technology Innovation Program of Hunan Province under Grant 2023RC1048, in part by the Hunan Provincial Natural Science Foundation of China under Grant 2024JJ3013.

\bibliographystyle{ACM-Reference-Format}
\balance
\bibliography{sample-base}
\end{document}